\pdfoutput=1

\documentclass{article}

\usepackage{microtype}
\usepackage{graphicx}
\usepackage{subfigure}
\usepackage{booktabs} %
\usepackage{xcolor}
\usepackage{multirow,makecell}
\usepackage{xspace}

\newcommand{\codeurl}{\url{https://github.com/machelreid/can-wikipedia-help-offline-rl}}
\newcommand{\chibit}{ChibiT\xspace}
\newcommand{\speedx}{3-6x}

\usepackage{hyperref}

\usepackage{amsmath,amsfonts,bm}

\def\eqref#1{equation~\ref{#1}}

\def\1{\bm{1}}

\def\rvt{{\mathbf{t}}}

\def\rvx{{\mathbf{x}}}

\DeclareMathAlphabet{\mathsfit}{\encodingdefault}{\sfdefault}{m}{sl}
\SetMathAlphabet{\mathsfit}{bold}{\encodingdefault}{\sfdefault}{bx}{n}

\def\gC{{\mathcal{C}}}

\def\gL{{\mathcal{L}}}

\usepackage[accepted]{icml2022}

\icmltitlerunning{Can Wikipedia Help Offline Reinforcement Learning?}

\begin{document}

\twocolumn[
\icmltitle{Can Wikipedia Help Offline Reinforcement Learning?}

\icmlsetsymbol{equal}{*}

\begin{icmlauthorlist}
\icmlauthor{Machel Reid}{utokyo}
\icmlauthor{Yutaro Yamada}{yale}
\icmlauthor{Shixiang Shane Gu}{goog}
\end{icmlauthorlist}

\icmlaffiliation{utokyo}{Graduate School of Engineering, The University of Tokyo}
\icmlaffiliation{goog}{Google Brain}
\icmlaffiliation{yale}{Department of Statistics and Data Science, Yale University}

\icmlcorrespondingauthor{Machel Reid}{machelreid@weblab.t.u-tokyo.ac.jp}

\icmlkeywords{Machine Learning, ICML}

\vskip 0.3in
]

\printAffiliationsAndNotice{\icmlEqualContribution} %

\begin{abstract}
Fine-tuning reinforcement learning (RL) models has been challenging because of a lack of large scale off-the-shelf datasets as well as high variance in transferability among different environments. Recent work has looked at tackling offline RL from the perspective of sequence modeling with improved results as result of the introduction of the Transformer architecture. However, when the model is trained from scratch, it suffers from slow convergence speeds. In this paper, we look to take advantage of this formulation of reinforcement learning as sequence modeling and investigate the transferability of pre-trained sequence models on other domains (vision, language) when finetuned on offline RL tasks (control, games). To this end, we also propose techniques to improve transfer between these domains. Results show consistent performance gains in terms of both convergence speed and reward on a variety of environments, accelerating training by \speedx~and achieving state-of-the-art performance in a variety of tasks using Wikipedia-pretrained and GPT2 language models. We hope that this work not only brings light to the potentials of leveraging generic sequence modeling techniques and pre-trained models for RL, but also inspires future work on sharing knowledge between generative modeling tasks of completely different domains.\footnote{We release code for reproducibility at \codeurl}

\end{abstract}
\begin{figure*}[]
	\centering
	\includegraphics[width=0.95\textwidth]{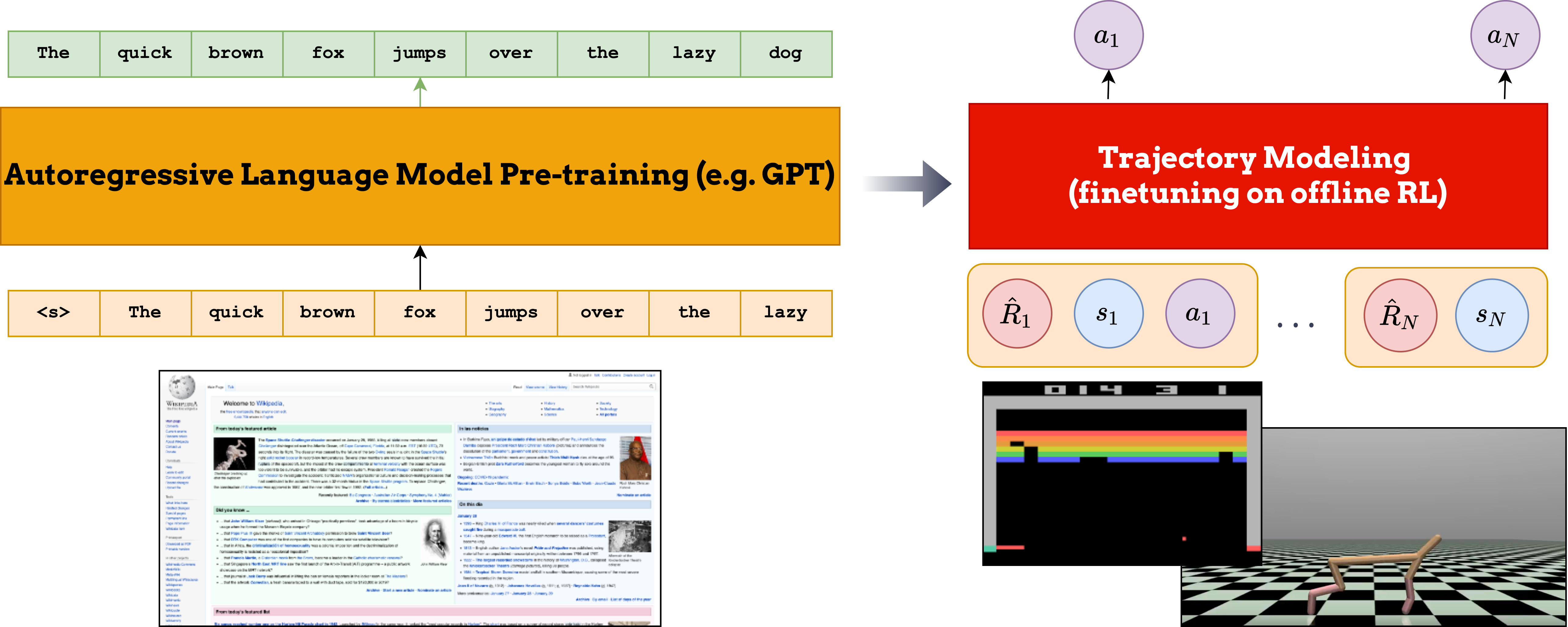} \label{fig:main}
	\caption{Adapting pre-trained language models (e.g. from Wikipedia) to offline RL (e.g. in continuous control and games).} %
\end{figure*}

\section{Introduction}
Large pre-trained language models have shown impressive performance in natural language \citep{devlin2019bert,Radford2018ImprovingLU} and vision \citep{dosovitskiy2021image} tasks. Furthermore, Transformer-based autoregressive language models \citep{vaswani2017attention,baevski2018adaptive,radford2019language} have shown to be powerful sources of zero-shot and few-shot performance \citep{brown2020language}, with notable rapid adaptation in low resource settings, demonstrating their easy adaptability and transferability to a number of tasks in their respective domains. Adapting autoregressive language models has also been extended to the multimodal setting \citep{tsimpoukelli2021multimodal} for tasks such as visual question answering. 

Concurrently, offline reinforcement learning (RL) has been seen as analogous to sequence modeling \citep{chen2021decision, janner2021reinforcement, furuta2021generalized}, framed as simply supervised learning to fit return-augmented trajectories in an offline dataset. This relaxation, doing away with many of the complexities commonly associated with reinforcement learning~\citep{watkins1992q,kakade2001natural}, allows us to take advantage of techniques popularized in sequence modeling tasks for RL. 

Pre-training, particularly, is an essential technique for alleviating higher compute costs from using more expressive models such as Transformers. However, such concept is still relatively fresh in RL~\citep{singh2020parrot,tirumala2020behavior}, due to the difficulty in parameterizing different scenes and tasks through a single network~\citep{wang2018nervenet,jiang2019language,zeng2020transporter} as well as the lack of large off-the-shelf datasets for pre-training~\citep{cobbe2020leveraging,zhu2020robosuite,yu2020meta}. Adopting pre-training as a default option for recent Transformer-based methods~\citep{chen2021decision, janner2021reinforcement, furuta2021generalized} appears far away -- if we only look within RL.

Unified under the umbrella of sequence modeling, we look at whether Transformer-based pre-trained \textit{language} models 
are able to be adapted to standard offline reinforcement learning tasks \textit{that have no relations to language}.  Given the setting of having a single model pre-trained on natural language to finetune on each offline RL task individually, we demonstrate drastic improvements in convergence speeds and final policy performances. We also consider further techniques (e.g. extension of positional embeddings, embedding similarity encouragement) in order to better take advantage of the features learned by the pre-trained language model and demonstrate greater improvements.

 We demonstrate that pre-training on autoregressively modeling natural language provides consistent performance gains when compared to the Decision Transformer \citep{chen2021decision} on both the popular OpenAI Gym \citep{brockman2016gym} and Atari \citep{bellemare13arcade} offline RL benchmarks. We also note a significantly faster convergence speed, with a \speedx~improvement over a vanilla Decision Transformer turning hours of training to tens of minutes, indicating long-term computational efficiency benefits on language pre-training.

Our findings allude to the potential impact of large scale pre-training for reinforcement learning, given its surprising efficacy when transferring from a  %
distant sequence modeling domain such as natural language. 
Notably, unlike other work on multi-task offline RL, our model provides consistent results in terms of both reward and convergence regardless of environment and setting, indicating a forseeable future where everyone should use a pre-trained language model for offline RL.

\section{Background}
\paragraph{Offline Reinforcement Learning}
We consider a standard Markov Decision Process (MDP) with state space $s\in \mathcal{S}$ and action space $a\in \mathcal{A}$, specified by a initial state distribution $p(s_1)$, a dynamics distribution $p(s_{t+1}|s_t,a_t)$, and a scalar reward function $r(s, a)$. The goal of reinforcement learning (RL) is to find the optimal policy $\pi^*(a|s)$ which maximizes the $\gamma$-discounted expected return as the agent interacts in the environment, 
\begin{align}
    \max_{\pi} \mathbb{E}_{s_{1:\infty}, a_{1:\infty}\sim p,\pi}\left[\sum_{t=1}^\infty \gamma^t r(s_t,a_t)\right]
\end{align}
In \textit{offline} RL, the objective remains the same, but has to be optimized with no interactive data collection on a fixed set of trajectories $\tau_i$, each of the form below with horizon $N$,
\begin{equation}
	\tau = (r_1, s_1, a_1, r_2, s_2, a_2, \dots, r_N, s_N, a_N).
\end{equation}
Common approaches include value-based or model-based objectives with regularization~\citep{fujimoto2019off,levine2020offline}, and more recently, direct generative modeling of these trajectories conditioned on hindsight returns~\citep{chen2021decision,janner2021reinforcement,furuta2021generalized}.  

\paragraph{Transformer model}
In this subsection, we briefly review the Transformer architecture \citep{vaswani2017attention} used to model sequences. The Transformer is comprised of stacks of identical \textit{Transformer layers}. Each of these layers takes in a set of $n$-dimensional vectors that are fed through the two main building blocks: a multi-head self-attention sublayer and a feedfoward MLP as shown below:
\begin{align}
	\text{Attention}(x) & = \text{softmax}\big(\frac{Q(x)K(x)^\top}{\sqrt{n}}\big)V(x) \\
	\text{Feedforward}(x) & = L_{2}(g(L_{1}(x)))
\end{align}
where $Q,K$ and $V$ represent linear projections  that parameterize the projection of input $x$ into the query, key and value spaces; while $L_1$, $L_2$ and $g$ represent the first linear projection,  second linear projection, and activation function that comprise the feedforward MLP. This is followed by a residual connection \citep{he2015deep} and layer normalization \citep{ba2016layer}.

\paragraph{Autoregressive Language Model Pre-training}
Although there are now multiple techniques for language model pre-training (e.g. masked language modeling; \citealp{devlin2019bert}), we will review autoregressive language modeling given its correspondence with the sequence modeling objective we employ for our offline reinforcement learning tasks.

Given a sequence $\rvx = [\rvx_1, \rvx_2, \dots \rvx_N]$ comprised of tokens $\rvx_i$, we look to model the likelihood of the sequence $P(\rvx)$ by way of modeling the probability of predicting each token $\rvx_i$ in a step-by-step, or autoregressive, fashion (commonly left-to-right). Naturally, it follows that each tokens prediction will be conditioned on all the previous elements in the sequence $\rvx_{<i}$ as shown below \citep{bengio2001neural}:

\begin{equation}
	P(\rvx) = \prod_{i=1}^{N} p(\rvx_i | \rvx_{i-1}, \rvx_{i-2}, \dots, \rvx_{1}) \label{eq:LM}
\end{equation}

\section{Methodology}
In this section we discuss our proposed methodology and techniques to better adapt pre-trained language models to model trajectories, as in the case of offline RL tasks with minimal modification to architecture and objectives shown in Figure~\ref{fig:main}.

\subsection{Modeling}
Following \citet{chen2021decision}, we model trajectories autoregressively by representing them in the following manner:
\begin{equation}
	\rvt = (\hat{R}_1, s_1, a_1, \hat{R}_2, s_2, a_2, \dots, \hat{R}_N, s_N, a_N)
\end{equation}
where trajectory $\rvt$ is modeled analogously to sequence $\rvx$ as shown in in Equation~\ref{eq:LM}, and $\hat{R}_i=\sum_{t=i}^N r_t,s_i, a_i$ represent the returns-to-go, state and action for each timestep $i$ given $N$ timesteps, respectively.
\subsection{Techniques}
\paragraph{Encouraging similarity between language representations and offline RL input representations} We find the issue of lack of alignment between state, action and reward input representations and language representations --- partially holding back further extraction of the capabilities of the language model. To this end, we use a similarity-based objective in order to maximize the similarity between the set of language embeddings
$E = [E_1,\dots,E_V]$ with vocabulary size $V$ and the set of input representations $I = I_1,\dots,I_{3N}$. The input representations are parameterized by linear projections $L_r, L_a, L_s$ corresponding to the target reward projection, action projection and state projection, respectively.

Given the following cosine similarity function:
\begin{align}
	\gC(z_1, z_2) &=  \frac{z_1}{\|{z_1}\|_2} \cdot \frac{z_2}{\|{z_2}\|_2}
\end{align}
we compute the negative (as we use gradient descent to optimize this objective) of the sum of the maximum similarity value for each embedding $E_1,\dots,E_j,\dots,E_V$ and each input representation $I_0,\dots,I_i,\dots,I_N$ as follows: \footnote{We looked at using mean pooling instead of max pooling for this objective and found that models with the mean pooling objective did not converge.} 
\begin{align}
	\gL_{\cos} = -\sum_{i=0}^{3N} \max_j \gC(I_i, E_j)
\end{align}
This allows us to encourage the input embeddings to become more similar to their language counterparts. However, due to computational cost of computing this loss for large values of $V$, we propose to use $K$-means clustering over the embeddings to reduce the size of $V$ to number of clusters $K$. We then treat the cluster centers akin to the original embeddings in order to compute our loss. Furthermore, we optimize this computation with vectorization.

\paragraph{Language model co-training} We also experiment with continuing to train jointly on language modeling and trajectory modeling. This allows us to encouraging the model's transformer backbone to be able to handle both language and trajectories simultaneously.

\subsection{Final Objective}
We now combine the objectives into the final objective below:
\begin{equation}
	\gL = \gL_{\text{MSE}} + \lambda_1 \gL_{\cos} + \lambda_2 \gL_{\text{LM}}
\end{equation}
where $\gL_{\text{MSE}}$ represents the mean squared error loss used for the primary trajectory modeling objective, $\gL_{\text{LM}}$ represents the negative log likelihood-based language modeling objective,  and $\lambda_1, \lambda_2$ represent hyperparameters to control the weight of the cosine similarity loss and language modeling loss, respectively.
\section{Experiments}
\begin{table*}[h]
\centering
\small
\begin{tabular}{lrrrrrrr}
\toprule
\multicolumn{1}{c}{\bf Game} & \multicolumn{1}{c}{\bf \chibit}& \multicolumn{1}{c}{\bf GPT2} & \multicolumn{1}{c}{\bf DT}& \multicolumn{1}{c}{\bf CQL} & \multicolumn{1}{c}{\bf QR-DQN} & \multicolumn{1}{c}{\bf REM} & \multicolumn{1}{c}{\bf BC} \\ %
\midrule
Breakout  &$280.3\pm63.7$& $\bf287.8\pm78.5$& ${267.5} $ & $211.1$ & $21.1$ & $32.1$ & $138.9$ \\ %
Qbert     &$ 22.3 \pm 9.3$ & $22.5\pm12.8$& $15.4$ & $\bf{104.2}$ & $1.7$ & $1.4$ & $17.3$ \\ %
Pong      &$\bf 112.3\pm7.2$& $111.0\pm5.7$ & $106.1$ & ${111.9}$ & $20.0$ & $39.1$ & $85.2$ \\ %
Seaquest  &$2.9 \pm 0.3$ & $\bf 3.0\pm0.2$ & $2.5$ & $1.7$ &  $1.4$ & $1.0$  & $2.1$ \\ %
\bottomrule
\end{tabular}
\caption{
Gamer-normalized scores for the 1\% DQN-replay Atari dataset.
We report the mean and variance across three seeds. Highest mean scores are highlighted in bold. }
\label{tbl:atari_main}
\end{table*}
\begin{table*}[h]
\centering
\footnotesize
\resizebox{0.98\textwidth}{!}{\begin{tabular}{llrrrrrrrrrrr}
\toprule
\multicolumn{1}{c}{\bf Dataset} &  \multicolumn{1}{c}{\bf Environment} &  \multicolumn{1}{c}{\bf \chibit} & \multicolumn{1}{c}{\bf GPT2} & \multicolumn{1}{c}{\bf CLIP} & \multicolumn{1}{c}{\bf iGPT} & \multicolumn{1}{c}{\bf DT} & \multicolumn{1}{c}{\bf CQL} & \multicolumn{1}{c}{\bf TD3+BC} & \multicolumn{1}{c}{\bf BRAC-v} & \multicolumn{1}{c}{\bf AWR} & \multicolumn{1}{c}{\bf BC} \\
\midrule
\multirowcell{3}{Medium\\Expert} & HalfCheetah &$\bf{91.7}\pm1.1$ & $\bf 91.8 \pm 0.5$& $91.3 \pm 0.4$ & $1.9 \pm 0.1$ & $86.8$ &  $62.4$ &  $90.7$ &  $41.9$ & $52.7$ & $59.9$ \\
& Hopper   &$\bf{110.0}\pm1.2$& $\bf110.9\pm1.6$& $110.2 \pm 0.1$ & $6.9 \pm 3.7$ & $107.6$ & $\bf{111.0}$ &  $98.0$ &   $0.8$ & $27.1$ & $79.6$ \\
 & Walker      & ${108.4}\pm0.2$  &${108.9} \pm 0.3$& $108.5 \pm 0.6$ & $0.5 \pm 0.7$ & $108.1$ &  $98.7$ &  $\bf 110.1$ &  $81.6$ & $53.8$ & $36.6$ \\
\midrule
\multirowcell{3}{{Medium}}        & HalfCheetah  &$43.3\pm0.1$ & $42.8\pm0.1$& $42.3 \pm 0.2$  & $1.5 \pm 0.1$ & $42.6$ &  $44.4$ &  $\bf 48.3$ &  $46.3$ & $37.4$ & $43.1$ \\
        & Hopper       &$\bf{82.1}\pm4.6$ &$79.1\pm1.1$& $66.9 \pm 0.9$ & $5.7 \pm 1.5$ & ${67.6}$ &   $58.0$ &  $59.3$ &  $31.1$ & $35.9$ & $63.9$ \\
        & Walker       &$77.8\pm0.1$& $78.3\pm1.5$& $74.1 \pm 0.9$ & $0.4 \pm 0.4$ &  $74.0$ &  $79.2$ &  $\bf 83.7$ & $81.1$ & $17.4$ & $77.3$ \\
\midrule
\multirowcell{3}{Medium\\Replay} & HalfCheetah &$39.7\pm0.5$& $40.3 \pm 2.3$& $37.9 \pm 0.2$ & $1.6 \pm 0.1$ & $36.6$ &  $46.2$ &  $44.6$ &  $\bf{47.7}$ & $40.3$ & $4.3$ \\
& Hopper      &$81.3\pm5.0$& $\bf 94.4\pm2.5$& $85.8 \pm 0.3$ & $5.7 \pm 0.9$ &  ${82.7} $ &  $48.6$ &  $60.9$ &   $0.6$ & $28.4$ & $27.6$ \\
& Walker      &$ 71.3\pm2.0$& $ 72.7\pm1.2$& $69.9 \pm 0.3$ & $9.1 \pm 7.7$ &  ${66.6} $ &  $26.7$ &  $\bf 81.8 $ &   $0.9$ & $15.5$ & $36.9$ \\
\midrule
\multicolumn{2}{c}{\bf Average (All Settings)}    &\bf 78.3 & \bf 80.1 & \bf 76.3 & 3.7 &  74.7 &63.9& 75.3 &36.9& 34.3 &46.4 \\
\bottomrule
\end{tabular}}
\caption{
Results for D4RL datasets\protect\footnotemark.
We report the mean and variance for three seeds. Language model pre-trainined models are consistently better than the Decision Transformer, and outperform/are competitive other baselines.
}
\label{tbl:mujoco_results}
\end{table*}

\subsection{Models} 

\paragraph{Pre-trained Models} We use the popular \textbf{GPT2}-small model to benchmark the impact of language-only pre-training. For direct comparison with the Decision Transformer \citep{chen2021decision}, we also pre-train a language model with the same parameter count on the popular language modeling Wikitext-103 dataset \citep{merity2016pointer}, consisting of over 100 million tokens from full Wikipedia articles. We refer to this model as \textbf{\chibit}.\footnote{``Chibi'' means ``small'' or ``mini'' in Japanese.} %

To explore the effect of pre-training on vision datasets, we also study \textbf{CLIP} \citep{Radford2021LearningTV} and \textbf{ImageGPT} \citep{chen20imagegpt}.
CLIP is comprised of an image encoder and a text encoder, and trained to predict which caption matches with which image. 
While the text encoder is an autoregressive Transformer, the image encoder is a Vision Transformer, which is not autoregressive. 
Therefore, for the autoregressive setup of offline reinforcement learning, we use the pre-trained text encoder as our initializer, while discarding the image encoder part.
ImageGPT is based on the same Transformer architecture as GPT2, but instead of language, it is trained on images unrolled into long sequences of pixels in an autoregressive manner.

\paragraph{RL Baselines}
In addition to benchmarking our pre-trained language models, we compare to popular state-of-the-art offline RL algorithms as follows: Decision Transformer (DT) \citep{chen2021decision}, CQL \citep{kumar2020conservative}, 
TD3+BC \citep{fujimoto2021minimalist},
BRAC \citep{wu2019behavior}, and AWR baselines \citep{peng2019advantageweighted}.

\paragraph{Hyperparameters} We use the following hyperparameters for our language model pre-training: the architecture is the same as that of \citet{chen2021decision} (128 model dim, 1 attention head, 3 layers), learning rate of $3\text{e-}4$, a batch size 65536 tokens, for 6 hours (80000 steps), using a warmup schedule over the first 10000. We the same byte-pair encoding (BPE; \citealp{sennrich-etal-2016-neural,kudo2018sentencepiece}) as that used by GPT-2 \citep{radford2019language}. %
For our offline RL tasks, we follow the hyperparameters used by \citep{chen2021decision}. For our additional objectives, we decay $\lambda_1,\lambda_2$, to reach $0.0$ each after 5000 steps.  We tune initial values of $\lambda_1$ for values of $\{0.1,0.2\}$ and $\lambda_2$ for values of $\{0.0,0.2,0.4\}$. We include additional details in the appendix. %

We benchmark our models against the D4RL offline RL benchmark datasets~\citep{fu2020d4rl} for the OpenAI Gym MuJoCo \citep{brockman2016gym} and Atari \citep{bellemare13arcade} tasks.

\subsection{Atari}
We run our \chibit and GPT2 models on the challenging Atari dataset \citep{bellemare13arcade}. We use the four Atari tasks evaluated in \citet{agarwal2020optimistic}, namely Breakout, Qbert, Pong and Seaquest.
Baseline numbers used are provided by \citet{chen2021decision} for behavior cloning and Decision Transformer models, while CQL, REM, and QR-QDN baseline numbers are provided by \citet{kumar2020conservative,agarwal2020optimistic}. Following \citet{hafner2021mastering}, we normalize scores based on that of a professional gamer on the evaluation set.

We show results in Table~\ref{tbl:atari_main}. It can be seen that \chibit and GPT2 %
results consistently improve over/match a strong vanilla Decision Transformer baseline. Our models are competitive with the Decision Transformer on all four games and competitive with CQL on 3/4 games.

\begin{figure*}[!htb]
	\centering
	\includegraphics[width=0.95\textwidth]{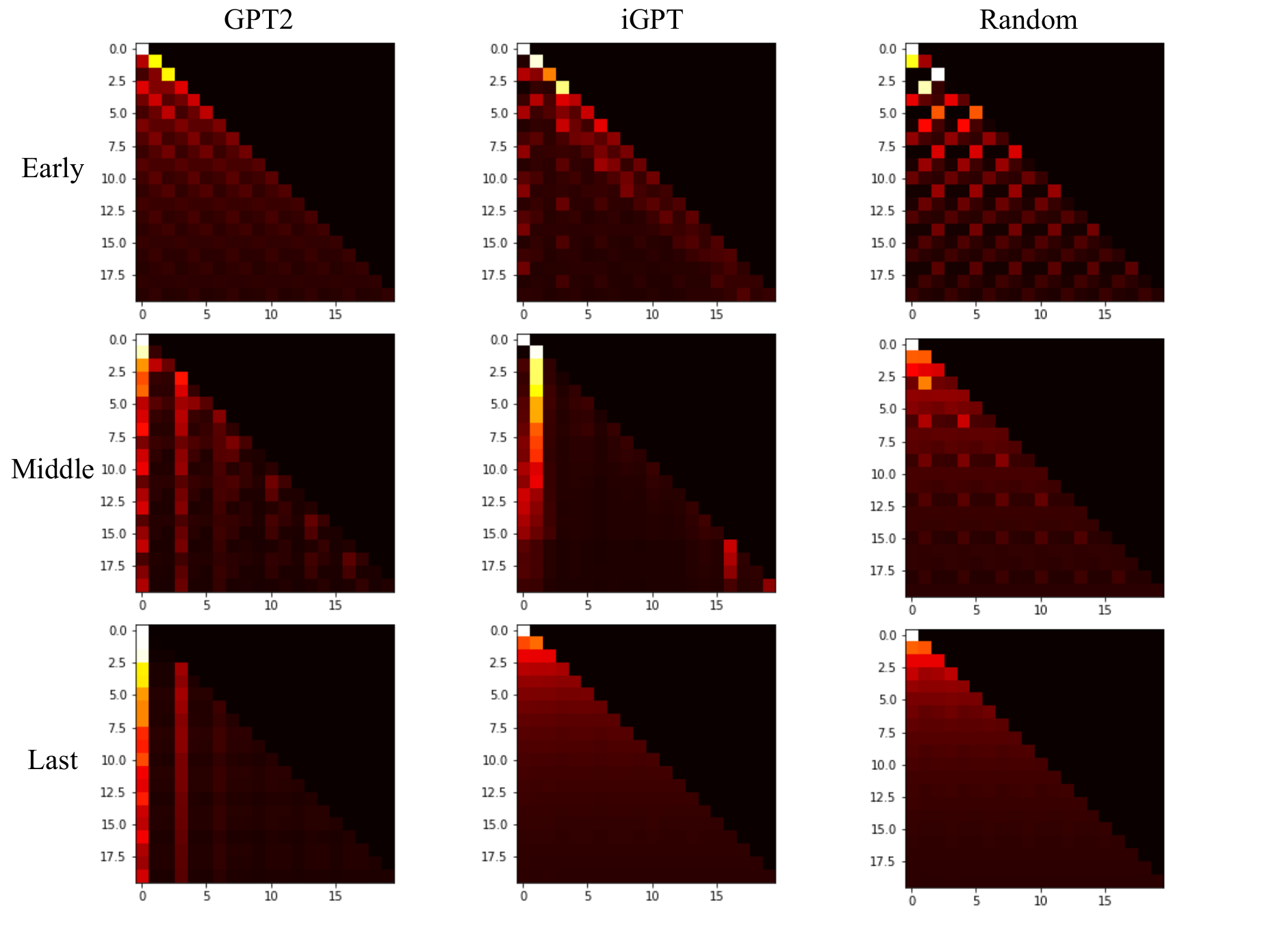} \label{fig:attention}
	\caption{\textbf{Attention analysis.} We visualize early, middle and last attention weights computed by GPT-2, iGPT, and randomly initialized DT models on Hopper-medium to study how pre-training on different modalities affects how the model attends to previous timesteps. The x-axis represents keys (representations that are being ``looked at'') while the y-axis represents queries (i.e. representations that are ``looking at'' other representations) for a given timestep.
	Ligher colors represent higher attention weights, while darker colors represent lower weights.}
\end{figure*}\subsection{Gym}
In this section, we consider results on the OpenAI Gym tasks (HalfCheetah, Walker2d, and Hopper) from the D4RL benchmark \citep{fu2020d4rl}.

We train our models for a total of 100k timesteps and evaluate every 5000 timesteps, with each evaluation consisting of 10 episodes. Note that we perform early stopping. Baseline results are obtained directly from the D4RL paper \citep{fu2020d4rl} and Decision Transformer results are directly taken from \citet{chen2021decision}. Similarly, following \citet{fu2020d4rl}, we compute the normalized score over returns, computed by taking $100\times\frac{\texttt{score-random score}}{\texttt{expert score - random score}}$. 

We show results comparing \chibit, GPT2, and CLIP with state-of-the-art offline RL algorithms in Table~\ref{tbl:mujoco_results}. Pre-training improves the Decision Transformer by large margins in an overwhelming majority of tasks, clearly demonstrating  that language pre-training improves over random initialization using sequence modeling techniques in terms of reward. We also take note of the minimal difference between \chibit, CLIP, and  GPT2, showing that that at this scale, improvements on offline RL are not necessarily strongly correlated with model size as has been shown on both large-scale vision and language tasks. We note that CLIP, while improving over a vanilla DT model, is often slightly less competitive that our pure language modeling objectives. Our \chibit and GPT2 models achieve and average performance of $78.3$ and $80.1$, respectively, showing strong competitiveness on all settings with all baselines. These pre-trained language models acheive state-of-the-art results by outperforming the strong Decision Transformer and TD3+BC baselines by a significant 3.0-5.4 points.

\section{Analysis}
In this section, we look at more fine-grained details and properties of various aspects of adapting pre-trained language models to offline RL tasks with ablations on OpenAI Gym.
\begin{table}[!htb]
    \centering
    \resizebox{0.48\textwidth}{!}{
    \begin{tabular}{lccc}
	    \toprule
	    \bf Model & \bf Walker2d & \bf HalfCheetah & \bf Hopper \\
	    \midrule 
	    DT (GitHub) & 3h14m & 3h23m & 2h47m \\
	    \chibit (ours) & 43m  & 48m  &36m  \\
	    GPT2 (ours) & 1h27m & 1h32m & 1h2m  \\
	    \bottomrule
    \end{tabular}}
    \caption{Training time comparison (measured in hours and minutes on a single V100 GPU on the medium-expert setting) between the Decision Transformer and two pre-trained models: \chibit and GPT2 on OpenAI gym tasks. Note that GPT2 is 144x larger than the other models with 84M model parameters.}
    \label{tab:convergence}
\end{table}
\subsection{Convergence Speed}
We evaluate time-to-convergence of GPT2, \chibit and DT using the our implementations of the former two and the author-provided implementation of the latter. Results are reported in Table~\ref{tab:convergence}. We find that pre-training on language allows us to speed up the training process of Transformer-based offline RL models, measured in wall-clock time. Convergence is defined as the point where average performance attains a score within 2 (normalized score) of the best score. %
Interestingly, we also find that GPT2, despite its larger model size at  84M model parameters, still manages to train faster than DT. This points towards potential benefits of pre-training %
at scale and increased efficiency during finetuning. We run experiments on a single NVIDIA V100 16GB GPU and an Intel Xeon Gold 6148 Processor.

\subsection{Language initialization versus vision initialization}

As we establish that Transformers pre-trained on language data are surprisingly effective for accelerating training convergence time on offline reinforcement learning tasks, it is tempting to ask if this phenomenon is inherent to language pre-training or does it extend to vision pre-training as well.
To answer this question, we compare two GPT models, ImageGPT-small (iGPT) and GPT2-small (GPT2), pre-trained on language and vision data, respectively. Since Transformer architectures are domain-agnostic, these models can be trained on 1D sequences of any form. Hence, we can compare GPT2, which was pre-trained on many sequences of discrete language tokens, and iGPT, which was pre-trained on autoregressive image generation at the pixel level (note that both models were trained on $\sim10^{10}$ tokens). %
Given the results in Table~\ref{tbl:mujoco_results} for iGPT, we found that the model had extremely low returns, and did not reach convergence. Notably, on some seeds, the model even performed worse than a random score after training on Walker medium, with a normalized score of $-0.1$, in contrast with GPT-2 pre-training which gives us an average increase of $5.1$ points  (measured in terms of normalized reward) over the Decision Transformer.

Furthermore, when we turn our attention to the difference between GPT2 and CLIP, we see that GPT2, which is based on pure-language based pre-training, performs better.
While the text encoder of CLIP is also an autoregressive Transformer pre-trained on text data, the objective of CLIP is different from GPT2 in that the former attempts to match the text description with their corresponding image, while the latter is pre-trained on pure autoregressive language modeling. Given this, we hypothesize that generative (versus discriminative) training objective is more useful for transfer to a generative task. %

We believe that this alludes to underlying similarities between language modeling and trajectory modeling, whereas a large difference between image modeling and trajectory modeling. Perhaps this can be attributed to the ``natural'' sequential nature of language and trajectories, versus the forced 2D$\rightarrow$1D nature that was used to pre-train iGPT.
\paragraph{Attention Analysis} 
To further understand the discrepancy between language-based and vision-based pre-training, we visualize attention weights, extracted from GPT2 and iGPT after fine-tuning on Hopper medium, as an example offline RL task. 
As a reference, we also extract attention weights from randomly initialized networks of Decision Transformers.
In Figure \ref{fig:attention}, we plot the attention weights averaged over all attention heads in each model, and present the visualizations for early, middle, and last layers, respectively.
Due to the autoregressive nature of our task, attention weights in the upper right triangle are masked out, so that the model can only attend to past sequences.

As a general trend, we see that in earlier layers GPT2 and the randomly initialized model tend to attend to positions with multiples of 3 timesteps behind the current position. This indicates that actions attend to previous actions, states attend to previous states, and returns-to-go attend to previous returns-to-go. Constrasted with this, iGPT's attention is less interpretable, however showing a notably stronger recency bias. In the middle layers, DT continues the trends of its early layers, whereas iGPT tends to fixate on a single state (given the overwhelming brightness of timestep $2$), GPT2 starts showing a stronger preference for previous returns to go (given that lighter colors are consistently timestep $1$, $4$, etc...). Finally, in the models' last layer, while iGPT and random initialization tend to exhibit a behaviour closer to mean pooling over all previous inputs, GPT's final prediction seems to be heavily reliant on the initial returns-to-go. This perhaps indicates that goal conditioning is stronger in GPT2. %

\subsection{How important is the model size of Transformer?}
We explore how pre-training changes the impact on model size for these offline RL tasks. We train randomly initialized models with various parameter counts (approx. 600K, 3M, 18M, 84M) as well as language-pre-trained models on WikiText-103 with the same parameter counts. Exact hyperparameters for this experiment are given in the Appendix.\footnote{Note that when pre-training language models with 600K, 3M, and 18M parameters, we control that our pre-training takes exactly 6 hours on 4 V100 GPUs.}

\begin{figure}[!htb]
	\includegraphics[width=0.48\textwidth]{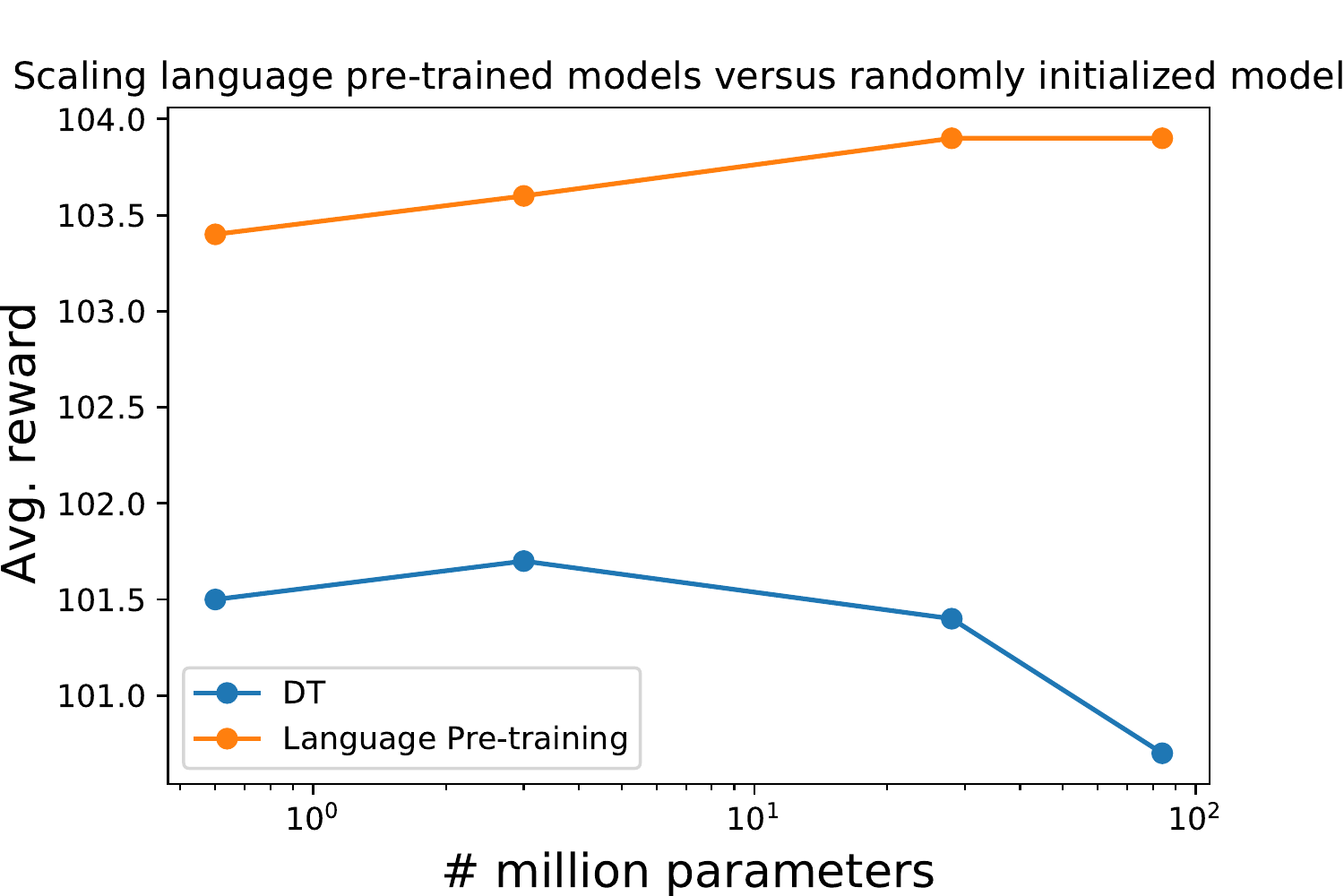} \vspace{-0.5cm} 
	\caption{Comparison of Average \textit{Medium-Expert} reward for various model sizes on OpenAI Gym.} \label{fig:params}
\end{figure}

We visualize the average (over Hopper, Walker2d, and HalfCheetah) of Medium-Expert results in Figure~\ref{fig:params}. Unsurprisingly, we observe that a randomly initialized Decision Transformer, tends to have lower relative returns as parameter sizes increase likely due to overfitting on finite data. Interestingly, however, pre-trained language models tend to increase performance as parameter count increases, despite diminishing returns with increasing parameter count. Nonetheless, this is exciting as it demonstrates that even language pre-training may be beneficial at scale, especially for larger and more diverse offline RL datasets in the future.  %

\begin{table}[]
    \centering
    \begin{tabular}{c|c}
	    \toprule
	    \bf Model & \bf Avg. Reward \\
	    \midrule
	    \chibit (context = 20) & 67.7 \\
	    \chibit (context = 60) & 67.3 \\
	    DT (context = 20) & 61.4 \\
	    DT (context = 60) & 61.2 \\
	    \bottomrule
    \end{tabular}
    \caption{Experiment on increased context length with pre-trained models on the medium setting}
    \label{tab:context}
\end{table}
\subsection{Context length}
We try various context lengths with pre-training and not pre-training: context = 20 (following \citet{chen2021decision}) and context = 60. Results are shown in Table~\ref{tab:context}. It can be seen that additional context does not seem to help even when pre-training on long range language modeling, perhaps alluding to the limited utility of long-range context for the OpenAI Gym tasks.
\subsection{Can we freeze model parameters?}
We also look at how \chibit performs when model weights (transformer blocks: self-attention and feedforward) are frozen with only action, state and return projections $L_a, L_s, L_r$ being trained. Previous work \citep{tsimpoukelli2021multimodal,lu2021pretrained} has demonstrated how frozen language models have the capability to extend to the vision domain with respectable performance, which we aim to test with this experiment. We show results on Table~\ref{tab:frozenft} on the D4RL medium setting in OpenAI Gym. When freezing model weights, performance is underwhelming with performance drastically reducing as much as $\sim$40\%. We conjecture this is due to our tasks being complex generative modeling as opposed to discriminative classification \citep{lu2021pretrained}, where the output distribution is of a higher dimension --- hence the need for more intensive finetuning. 
\begin{table}[!htb]
    \centering
    \footnotesize
    \begin{tabular}{lccc}
    \toprule
    \bf Model &\bf HalfCheetah &  \bf Walker2d & \bf Hopper \\

    \midrule 

    \chibit (FT) &    $43.3\pm0.1$ &$77.8\pm0.1$&${82.1}\pm4.6$ \\
    \chibit (Frozen)  &$26.4\pm1.2$  & $63.3\pm2.7$ & $57.7\pm7.0$ \\ 
    \bottomrule
    \end{tabular}         
    \caption{Experiment on freezing model weights versus finetuning them on OpenAI Gym.}
    \label{tab:frozenft}
\end{table}
\subsection{Ablation of proposed techniques}
We perform an ablation study of our proposed auxiliary techniques and compare the impact of including and not including pre-trained positional embeddings. Results are shown in Table~\ref{tab:ablation}. It can be seen that the combination of our objectives are able to increase performance consistently. We also note that the removal of pre-trained positional embeddings results in the largest average decrease in performance over \chibit, alluding to the fact that this positional information is important and transferable to offline RL.
\begin{table}[!htb]
    \centering
    \resizebox{0.48\textwidth}{!}{
    \begin{tabular}{lccc}
    \toprule
    \bf Model &\bf HalfCheetah &  \bf Walker2d & \bf Hopper \\
    \midrule 
    \chibit &    $43.3\pm0.1$    &$77.8\pm0.1$&${82.1}\pm4.6$ \\
    \chibit (w/o $\gL_{\cos}$)   & $43.1\pm0.1$ & $77.2\pm1.3$ & $80.9\pm1.1$  \\ 
    \chibit (w/o $\gL_{\text{LM}}$) & $43.3\pm0.2$ & $77.6\pm0.2$ & $81.4\pm5.2$\\ 
    \chibit (rand. pos. emb.)     &$43.0\pm0.4$  & $76.5\pm1.2$ & $78.4\pm2.0$ \\ 
    \bottomrule
    \end{tabular}}
    \caption{Ablation of our proposed techniques}
    \label{tab:ablation}
\end{table}

\section{Related Work}
\paragraph{Transformer Pre-training} 
Pre-training Transformer-based models \citep{vaswani2017attention} was initially proposed by \citet{Radford2018ImprovingLU} with their Generative Pre-trained Transformer (GPT). They performed autoregressive language modeling on a relatively large dataset, showing promising initial success not only on its ability to scale to large models sizes, but also for its impressive performance when fine-tuning on task-specific natural language understanding (NLU; \citealp{wang2018glue}) datasets. BERT \citep{devlin2019bert}, extended this pre-train$\rightarrow$finetune paradigm with their masked language modeling objective for pre-training which allowed the model to take advantage of its bidirectional attention capabilities for NLU tasks. Furthermore, recently this paradigm has extended to computer vision with the Vision Transformer (ViT; \citealp{dosovitskiy2021image}).
SwinTransformer \citep{liu2021swin}  extends ViT by introducing hierarchical multi-resolution feature maps. 
By pre-training SwinTransformer on ImageNet-22k, and fine-tuning on downstream tasks such as object detection and semantic segmentation, SwinTransfomer outperform previous state-of-the-arts models based on Convolutional Neural Networks (CNN)~\citep{su2020vlbert}.

\paragraph{Sequence Modeling for Offline RL} Offline RL became popular starting from a simple observation that many performant off-policy algorithms~\citep{mnih2015human,lillicrap2015continuous,gu2016continuous,haarnoja2018soft,fujimoto2018addressing} fail to learn in a fully off-policy, i.e. \textit{offline}, batch setting~\citep{fujimoto2019off}. Numerous algorithmic work ensued~\citep{wu2019behavior,jaques2020human,ghasemipour2021emaq,kumar2020conservative,fujimoto2021minimalist} with various applications~\citep{jaques2020human,chebotar2021actionable}. Building on reward-conditioned imitation learning~\citep{srivastava2019training,kumar2019reward}, Transformer architecture has been recently adopted for replacing offline RL with sequence modeling~\citep{chen2021decision,janner2021reinforcement,furuta2021generalized}. Despite initial successes, many techniques popular in language modeling have yet to be experimented in these offline RL benchmarks, and our work constitutes an initial step toward bridging the two communities.  

\paragraph{Pre-training for RL} Contrary to language or vision~\citep{devlin2019bert,dosovitskiy2021image}, major successes in deep RL have largely focused on isolated tasks or domains~\citep{mnih2015human,silver2016mastering,gu2017deep,kalashnikov2018qt,vinyals2019grandmaster}. Pre-training results are often limited to vision or language processing~\citep{yen2020learning,lynch2021language} or specially-crafted domains~\citep{singh2020parrot,tirumala2020behavior}. Arguably, a fundamental bottleneck for pre-training in RL is the difficulty in reusing a single network across vastly different tasks, of distinct observation spaces, action spaces, rewards, scenes, and agent morphologies. Preliminary work explored various aspects of this problem through graph neural networks for morphology generalization~\citep{wang2018nervenet,pathak2019learning,chen2018hardware,kurin2020my}, language for universal reward specification~\citep{jiang2019language,lynch2021language,shridhar2022cliport}, and object-centric action spaces~\citep{zeng2020transporter,shridhar2022cliport,noguchi2021tool}. Our work is orthogonal to these as we essentially amortize RL algorithm itself, expressed as sequence modeling with Transformer, instead of specific RL domain information, and can be combined with domain-specific pre-training techniques~\citep{yen2020learning,lynch2021language} effortlessly.

\paragraph{Adapting language models to new modalities and domains}
Within language modeling recently there has been interest in domain adaptation of pre-trained language models \citep{gururangan2020dont}, where it has been shown that continued modeling on a domain-specific datasets tends to lead to greater performance on domain-related downstream tasks. Furthermore, \citet{tsimpoukelli2021multimodal} looked at adapting frozen autoregressive language models for few-shot question answering by adding an auxiliary vision encoder. More related to our work is that of \citet{lu2021pretrained}, where they look at adapting frozen language models to various tasks such as image classification. Our work extends on the spirit of these works by adapting language models to a new domain of RL, however, as far was we know, we are the first to propose leveraging a generative model (in language) for generation in another domain (RL) as opposed to a discriminatory task such as classification.

\section{Conclusion}

We investigate how pre-trained models can improve generic offline RL problems, recently casted as sequence modeling. To our surprise, we discover that fine-tuning from a Wikipedia-trained small transformer (\chibit) or a GPT2 model outperforms the basic Decision Transformer (DT) and other RL-based offline baselines by a large margin in terms of policy performance and convergence, establishing state-of-the-art scores on the competitive D4RL benchmark in both Gym and Atari and cutting down the DT training time by \speedx. We perform extensive ablation studies and analyses, and found how language pre-training (as opposed to vision pre-training), model size, and fine-tuning (as opposed to freezing parameters) play critical roles in the final performances. We hope our work can accelerate the adoption of pre-training in RL and leads to more interest in applying other sequence modeling techniques from language and vision into RL.

Beyond RL, our work constitutes the first successful transfer, to the best of our knowledge, of a pre-trained generative model in one domain (language) to a generative modeling task in a completely different domain (RL on continuous control and games). This hints at some underlying universal structure across sequence modeling domains, and could perhaps lead to unified generative modeling pre-training for better transferability among them. %
In future work, we look to investigate in more depth which properties of language structure are useful for reinforcement learning and sequence modeling in other domains, and whether previous work studying language structure \citep{hupkes2019compositionality} does indeed relate to compositional generalization of neural networks.

\section*{Acknowledgements}
We thank Hiroki Furuta, Yusuke Iwasawa, Edison Marrese-Taylor  Graham Neubig, and Alfredo Solano for their comments. MR and YY also thank the Masason Foundation for their support.

\bibliography{example_paper}

\begin{thebibliography}{69}
\providecommand{\natexlab}[1]{#1}
\providecommand{\url}[1]{\texttt{#1}}
\expandafter\ifx\csname urlstyle\endcsname\relax
  \providecommand{\doi}[1]{doi: #1}\else
  \providecommand{\doi}{doi: \begingroup \urlstyle{rm}\Url}\fi

\bibitem[Agarwal et~al.(2020)Agarwal, Schuurmans, and
  Norouzi]{agarwal2020optimistic}
Agarwal, R., Schuurmans, D., and Norouzi, M.
\newblock An optimistic perspective on offline reinforcement learning, 2020.

\bibitem[Agarwal et~al.(2021)Agarwal, Schwarzer, Castro, Courville, and
  Bellemare]{DBLP:conf/nips/AgarwalSCCB21}
Agarwal, R., Schwarzer, M., Castro, P.~S., Courville, A.~C., and Bellemare,
  M.~G.
\newblock Deep reinforcement learning at the edge of the statistical precipice.
\newblock In Ranzato, M., Beygelzimer, A., Dauphin, Y.~N., Liang, P., and
  Vaughan, J.~W. (eds.), \emph{Advances in Neural Information Processing
  Systems 34: Annual Conference on Neural Information Processing Systems 2021,
  NeurIPS 2021, December 6-14, 2021, virtual}, pp.\  29304--29320, 2021.
\newblock URL
  \url{https://proceedings.neurips.cc/paper/2021/hash/f514cec81cb148559cf475e7426eed5e-Abstract.html}.

\bibitem[Ba et~al.(2016)Ba, Kiros, and Hinton]{ba2016layer}
Ba, J.~L., Kiros, J.~R., and Hinton, G.~E.
\newblock Layer normalization.
\newblock \emph{arXiv preprint arXiv: Arxiv-1607.06450}, 2016.

\bibitem[Baevski \& Auli(2019)Baevski and Auli]{baevski2018adaptive}
Baevski, A. and Auli, M.
\newblock Adaptive input representations for neural language modeling.
\newblock In \emph{International Conference on Learning Representations}, 2019.
\newblock URL \url{https://openreview.net/forum?id=ByxZX20qFQ}.

\bibitem[{Bellemare} et~al.(2013){Bellemare}, {Naddaf}, {Veness}, and
  {Bowling}]{bellemare13arcade}
{Bellemare}, M.~G., {Naddaf}, Y., {Veness}, J., and {Bowling}, M.
\newblock The arcade learning environment: An evaluation platform for general
  agents.
\newblock \emph{Journal of Artificial Intelligence Research}, 47:\penalty0
  253--279, jun 2013.

\bibitem[Bengio et~al.(2001)Bengio, Ducharme, and Vincent]{bengio2001neural}
Bengio, Y., Ducharme, R., and Vincent, P.
\newblock A neural probabilistic language model.
\newblock In Leen, T., Dietterich, T., and Tresp, V. (eds.), \emph{Advances in
  Neural Information Processing Systems}, volume~13. MIT Press, 2001.
\newblock URL
  \url{https://proceedings.neurips.cc/paper/2000/file/728f206c2a01bf572b5940d7d9a8fa4c-Paper.pdf}.

\bibitem[Brockman et~al.(2016)Brockman, Cheung, Pettersson, Schneider,
  Schulman, Tang, and Zaremba]{brockman2016gym}
Brockman, G., Cheung, V., Pettersson, L., Schneider, J., Schulman, J., Tang,
  J., and Zaremba, W.
\newblock Openai gym, 2016.

\bibitem[Brown et~al.(2020)Brown, Mann, Ryder, Subbiah, Kaplan, Dhariwal,
  Neelakantan, Shyam, Sastry, Askell, Agarwal, Herbert-Voss, Krueger, Henighan,
  Child, Ramesh, Ziegler, Wu, Winter, Hesse, Chen, Sigler, Litwin, Gray, Chess,
  Clark, Berner, McCandlish, Radford, Sutskever, and Amodei]{brown2020language}
Brown, T.~B., Mann, B., Ryder, N., Subbiah, M., Kaplan, J., Dhariwal, P.,
  Neelakantan, A., Shyam, P., Sastry, G., Askell, A., Agarwal, S.,
  Herbert-Voss, A., Krueger, G., Henighan, T., Child, R., Ramesh, A., Ziegler,
  D.~M., Wu, J., Winter, C., Hesse, C., Chen, M., Sigler, E., Litwin, M., Gray,
  S., Chess, B., Clark, J., Berner, C., McCandlish, S., Radford, A., Sutskever,
  I., and Amodei, D.
\newblock Language models are few-shot learners, 2020.

\bibitem[Chebotar et~al.(2021)Chebotar, Hausman, Lu, Xiao, Kalashnikov, Varley,
  Irpan, Eysenbach, Julian, Finn, et~al.]{chebotar2021actionable}
Chebotar, Y., Hausman, K., Lu, Y., Xiao, T., Kalashnikov, D., Varley, J.,
  Irpan, A., Eysenbach, B., Julian, R., Finn, C., et~al.
\newblock Actionable models: Unsupervised offline reinforcement learning of
  robotic skills.
\newblock \emph{arXiv preprint arXiv:2104.07749}, 2021.

\bibitem[Chen et~al.(2021)Chen, Lu, Rajeswaran, Lee, Grover, Laskin, Abbeel,
  Srinivas, and Mordatch]{chen2021decision}
Chen, L., Lu, K., Rajeswaran, A., Lee, K., Grover, A., Laskin, M., Abbeel, P.,
  Srinivas, A., and Mordatch, I.
\newblock Decision transformer: Reinforcement learning via sequence modeling.
\newblock \emph{arXiv preprint arXiv: Arxiv-2106.01345}, 2021.

\bibitem[Chen et~al.(2020)Chen, Radford, Child, Wu, Jun, Luan, and
  Sutskever]{chen20imagegpt}
Chen, M., Radford, A., Child, R., Wu, J., Jun, H., Luan, D., and Sutskever, I.
\newblock Generative pretraining from pixels.
\newblock In III, H.~D. and Singh, A. (eds.), \emph{Proceedings of the 37th
  International Conference on Machine Learning}, volume 119 of
  \emph{Proceedings of Machine Learning Research}, pp.\  1691--1703. PMLR,
  13--18 Jul 2020.
\newblock URL \url{https://proceedings.mlr.press/v119/chen20s.html}.

\bibitem[Chen et~al.(2018)Chen, Murali, and Gupta]{chen2018hardware}
Chen, T., Murali, A., and Gupta, A.
\newblock Hardware conditioned policies for multi-robot transfer learning.
\newblock \emph{arXiv preprint arXiv:1811.09864}, 2018.

\bibitem[Cobbe et~al.(2020)Cobbe, Hesse, Hilton, and
  Schulman]{cobbe2020leveraging}
Cobbe, K., Hesse, C., Hilton, J., and Schulman, J.
\newblock Leveraging procedural generation to benchmark reinforcement learning.
\newblock In \emph{International conference on machine learning}, pp.\
  2048--2056. PMLR, 2020.

\bibitem[Devlin et~al.(2019)Devlin, Chang, Lee, and Toutanova]{devlin2019bert}
Devlin, J., Chang, M.-W., Lee, K., and Toutanova, K.
\newblock Bert: Pre-training of deep bidirectional transformers for language
  understanding, 2019.

\bibitem[Dosovitskiy et~al.(2021)Dosovitskiy, Beyer, Kolesnikov, Weissenborn,
  Zhai, Unterthiner, Dehghani, Minderer, Heigold, Gelly, Uszkoreit, and
  Houlsby]{dosovitskiy2021image}
Dosovitskiy, A., Beyer, L., Kolesnikov, A., Weissenborn, D., Zhai, X.,
  Unterthiner, T., Dehghani, M., Minderer, M., Heigold, G., Gelly, S.,
  Uszkoreit, J., and Houlsby, N.
\newblock An image is worth 16x16 words: Transformers for image recognition at
  scale, 2021.

\bibitem[Fu et~al.(2020)Fu, Kumar, Nachum, Tucker, and Levine]{fu2020d4rl}
Fu, J., Kumar, A., Nachum, O., Tucker, G., and Levine, S.
\newblock D4rl: Datasets for deep data-driven reinforcement learning.
\newblock \emph{arXiv preprint arXiv: Arxiv-2004.07219}, 2020.

\bibitem[Fujimoto \& Gu(2021)Fujimoto and Gu]{fujimoto2021minimalist}
Fujimoto, S. and Gu, S.~S.
\newblock A minimalist approach to offline reinforcement learning.
\newblock \emph{arXiv preprint arXiv:2106.06860}, 2021.

\bibitem[Fujimoto et~al.(2018)Fujimoto, Hoof, and
  Meger]{fujimoto2018addressing}
Fujimoto, S., Hoof, H., and Meger, D.
\newblock Addressing function approximation error in actor-critic methods.
\newblock In \emph{International Conference on Machine Learning}, pp.\
  1587--1596. PMLR, 2018.

\bibitem[Fujimoto et~al.(2019)Fujimoto, Meger, and Precup]{fujimoto2019off}
Fujimoto, S., Meger, D., and Precup, D.
\newblock Off-policy deep reinforcement learning without exploration.
\newblock In \emph{International Conference on Machine Learning}, pp.\
  2052--2062. PMLR, 2019.

\bibitem[Furuta et~al.(2021)Furuta, Matsuo, and Gu]{furuta2021generalized}
Furuta, H., Matsuo, Y., and Gu, S.~S.
\newblock Generalized decision transformer for offline hindsight information
  matching.
\newblock \emph{arXiv preprint arXiv:2111.10364}, 2021.

\bibitem[Ghasemipour et~al.(2021)Ghasemipour, Schuurmans, and
  Gu]{ghasemipour2021emaq}
Ghasemipour, S. K.~S., Schuurmans, D., and Gu, S.~S.
\newblock Emaq: Expected-max q-learning operator for simple yet effective
  offline and online rl.
\newblock In \emph{International Conference on Machine Learning}, pp.\
  3682--3691. PMLR, 2021.

\bibitem[Gu et~al.(2016)Gu, Lillicrap, Sutskever, and Levine]{gu2016continuous}
Gu, S., Lillicrap, T., Sutskever, I., and Levine, S.
\newblock Continuous deep q-learning with model-based acceleration.
\newblock In \emph{International conference on machine learning}, pp.\
  2829--2838. PMLR, 2016.

\bibitem[Gu et~al.(2017)Gu, Holly, Lillicrap, and Levine]{gu2017deep}
Gu, S., Holly, E., Lillicrap, T., and Levine, S.
\newblock Deep reinforcement learning for robotic manipulation with
  asynchronous off-policy updates.
\newblock In \emph{2017 IEEE international conference on robotics and
  automation (ICRA)}, pp.\  3389--3396. IEEE, 2017.

\bibitem[Gururangan et~al.(2020)Gururangan, Marasović, Swayamdipta, Lo,
  Beltagy, Downey, and Smith]{gururangan2020dont}
Gururangan, S., Marasović, A., Swayamdipta, S., Lo, K., Beltagy, I., Downey,
  D., and Smith, N.~A.
\newblock Don't stop pretraining: Adapt language models to domains and tasks,
  2020.

\bibitem[Haarnoja et~al.(2018)Haarnoja, Zhou, Abbeel, and
  Levine]{haarnoja2018soft}
Haarnoja, T., Zhou, A., Abbeel, P., and Levine, S.
\newblock Soft actor-critic: Off-policy maximum entropy deep reinforcement
  learning with a stochastic actor.
\newblock In \emph{International conference on machine learning}, pp.\
  1861--1870. PMLR, 2018.

\bibitem[Hafner et~al.(2021)Hafner, Lillicrap, Norouzi, and
  Ba]{hafner2021mastering}
Hafner, D., Lillicrap, T., Norouzi, M., and Ba, J.
\newblock Mastering atari with discrete world models, 2021.

\bibitem[He et~al.(2015)He, Zhang, Ren, and Sun]{he2015deep}
He, K., Zhang, X., Ren, S., and Sun, J.
\newblock Deep residual learning for image recognition.
\newblock \emph{arXiv preprint arXiv: Arxiv-1512.03385}, 2015.

\bibitem[Hupkes et~al.(2019)Hupkes, Dankers, Mul, and
  Bruni]{hupkes2019compositionality}
Hupkes, D., Dankers, V., Mul, M., and Bruni, E.
\newblock Compositionality decomposed: how do neural networks generalise?
\newblock \emph{arXiv preprint arXiv: Arxiv-1908.08351}, 2019.

\bibitem[Janner et~al.(2021)Janner, Li, and Levine]{janner2021reinforcement}
Janner, M., Li, Q., and Levine, S.
\newblock Reinforcement learning as one big sequence modeling problem.
\newblock \emph{arXiv preprint arXiv:2106.02039}, 2021.

\bibitem[Jaques et~al.(2020)Jaques, Shen, Ghandeharioun, Ferguson, Lapedriza,
  Jones, Gu, and Picard]{jaques2020human}
Jaques, N., Shen, J.~H., Ghandeharioun, A., Ferguson, C., Lapedriza, A., Jones,
  N., Gu, S.~S., and Picard, R.
\newblock Human-centric dialog training via offline reinforcement learning.
\newblock \emph{arXiv preprint arXiv:2010.05848}, 2020.

\bibitem[Jiang et~al.(2019)Jiang, Gu, Murphy, and Finn]{jiang2019language}
Jiang, Y., Gu, S., Murphy, K., and Finn, C.
\newblock Language as an abstraction for hierarchical deep reinforcement
  learning.
\newblock \emph{arXiv preprint arXiv:1906.07343}, 2019.

\bibitem[Kakade(2001)]{kakade2001natural}
Kakade, S.~M.
\newblock A natural policy gradient.
\newblock \emph{Advances in neural information processing systems}, 14, 2001.

\bibitem[Kalashnikov et~al.(2018)Kalashnikov, Irpan, Pastor, Ibarz, Herzog,
  Jang, Quillen, Holly, Kalakrishnan, Vanhoucke, et~al.]{kalashnikov2018qt}
Kalashnikov, D., Irpan, A., Pastor, P., Ibarz, J., Herzog, A., Jang, E.,
  Quillen, D., Holly, E., Kalakrishnan, M., Vanhoucke, V., et~al.
\newblock Qt-opt: Scalable deep reinforcement learning for vision-based robotic
  manipulation.
\newblock \emph{arXiv preprint arXiv:1806.10293}, 2018.

\bibitem[Kudo \& Richardson(2018)Kudo and Richardson]{kudo2018sentencepiece}
Kudo, T. and Richardson, J.
\newblock Sentencepiece: A simple and language independent subword tokenizer
  and detokenizer for neural text processing, 2018.

\bibitem[Kumar et~al.(2019)Kumar, Peng, and Levine]{kumar2019reward}
Kumar, A., Peng, X.~B., and Levine, S.
\newblock Reward-conditioned policies.
\newblock \emph{arXiv preprint arXiv:1912.13465}, 2019.

\bibitem[Kumar et~al.(2020)Kumar, Zhou, Tucker, and
  Levine]{kumar2020conservative}
Kumar, A., Zhou, A., Tucker, G., and Levine, S.
\newblock Conservative q-learning for offline reinforcement learning.
\newblock \emph{arXiv preprint arXiv:2006.04779}, 2020.

\bibitem[Kurin et~al.(2020)Kurin, Igl, Rockt{\"a}schel, Boehmer, and
  Whiteson]{kurin2020my}
Kurin, V., Igl, M., Rockt{\"a}schel, T., Boehmer, W., and Whiteson, S.
\newblock My body is a cage: the role of morphology in graph-based incompatible
  control.
\newblock \emph{arXiv preprint arXiv:2010.01856}, 2020.

\bibitem[Levine et~al.(2020)Levine, Kumar, Tucker, and Fu]{levine2020offline}
Levine, S., Kumar, A., Tucker, G., and Fu, J.
\newblock Offline reinforcement learning: Tutorial, review, and perspectives on
  open problems.
\newblock \emph{arXiv preprint arXiv:2005.01643}, 2020.

\bibitem[Lillicrap et~al.(2015)Lillicrap, Hunt, Pritzel, Heess, Erez, Tassa,
  Silver, and Wierstra]{lillicrap2015continuous}
Lillicrap, T.~P., Hunt, J.~J., Pritzel, A., Heess, N., Erez, T., Tassa, Y.,
  Silver, D., and Wierstra, D.
\newblock Continuous control with deep reinforcement learning.
\newblock \emph{arXiv preprint arXiv:1509.02971}, 2015.

\bibitem[Liu et~al.(2021)Liu, Lin, Cao, Hu, Wei, Zhang, Lin, and
  Guo]{liu2021swin}
Liu, Z., Lin, Y., Cao, Y., Hu, H., Wei, Y., Zhang, Z., Lin, S., and Guo, B.
\newblock Swin transformer: Hierarchical vision transformer using shifted
  windows, 2021.

\bibitem[Lu et~al.(2021)Lu, Grover, Abbeel, and Mordatch]{lu2021pretrained}
Lu, K., Grover, A., Abbeel, P., and Mordatch, I.
\newblock Pretrained transformers as universal computation engines, 2021.

\bibitem[Lynch \& Sermanet(2021)Lynch and Sermanet]{lynch2021language}
Lynch, C. and Sermanet, P.
\newblock Language conditioned imitation learning over unstructured data.
\newblock \emph{Proceedings of Robotics: Science and Systems. doi}, 10, 2021.

\bibitem[Merity et~al.(2016)Merity, Xiong, Bradbury, and
  Socher]{merity2016pointer}
Merity, S., Xiong, C., Bradbury, J., and Socher, R.
\newblock Pointer sentinel mixture models, 2016.

\bibitem[Mnih et~al.(2015)Mnih, Kavukcuoglu, Silver, Rusu, Veness, Bellemare,
  Graves, Riedmiller, Fidjeland, Ostrovski, et~al.]{mnih2015human}
Mnih, V., Kavukcuoglu, K., Silver, D., Rusu, A.~A., Veness, J., Bellemare,
  M.~G., Graves, A., Riedmiller, M., Fidjeland, A.~K., Ostrovski, G., et~al.
\newblock Human-level control through deep reinforcement learning.
\newblock \emph{nature}, 518\penalty0 (7540):\penalty0 529--533, 2015.

\bibitem[Noguchi et~al.(2021)Noguchi, Matsushima, Matsuo, and
  Gu]{noguchi2021tool}
Noguchi, Y., Matsushima, T., Matsuo, Y., and Gu, S.~S.
\newblock Tool as embodiment for recursive manipulation.
\newblock \emph{arXiv preprint arXiv:2112.00359}, 2021.

\bibitem[Pathak et~al.(2019)Pathak, Lu, Darrell, Isola, and
  Efros]{pathak2019learning}
Pathak, D., Lu, C., Darrell, T., Isola, P., and Efros, A.~A.
\newblock Learning to control self-assembling morphologies: a study of
  generalization via modularity.
\newblock \emph{arXiv preprint arXiv:1902.05546}, 2019.

\bibitem[Peng et~al.(2019)Peng, Kumar, Zhang, and
  Levine]{peng2019advantageweighted}
Peng, X.~B., Kumar, A., Zhang, G., and Levine, S.
\newblock Advantage-weighted regression: Simple and scalable off-policy
  reinforcement learning.
\newblock \emph{arXiv preprint arXiv: Arxiv-1910.00177}, 2019.

\bibitem[Radford et~al.(2018)Radford, Narasimhan, Salimans, and
  Suskeveter]{Radford2018ImprovingLU}
Radford, A., Narasimhan, K., Salimans, T., and Suskeveter, I.
\newblock Improving language understanding by generative pre-training.
\newblock 2018.

\bibitem[Radford et~al.(2019)Radford, Wu, Child, Luan, Amodei, and
  Sutskever]{radford2019language}
Radford, A., Wu, J., Child, R., Luan, D., Amodei, D., and Sutskever, I.
\newblock Language models are unsupervised multitask learners.
\newblock 2019.

\bibitem[Radford et~al.(2021)Radford, Kim, Hallacy, Ramesh, Goh, Agarwal,
  Sastry, Askell, Mishkin, Clark, Krueger, and
  Sutskever]{Radford2021LearningTV}
Radford, A., Kim, J.~W., Hallacy, C., Ramesh, A., Goh, G., Agarwal, S., Sastry,
  G., Askell, A., Mishkin, P., Clark, J., Krueger, G., and Sutskever, I.
\newblock Learning transferable visual models from natural language
  supervision.
\newblock In \emph{ICML}, 2021.

\bibitem[Sennrich et~al.(2016)Sennrich, Haddow, and
  Birch]{sennrich-etal-2016-neural}
Sennrich, R., Haddow, B., and Birch, A.
\newblock Neural machine translation of rare words with subword units.
\newblock In \emph{Proceedings of the 54th Annual Meeting of the Association
  for Computational Linguistics (Volume 1: Long Papers)}, pp.\  1715--1725,
  Berlin, Germany, August 2016. Association for Computational Linguistics.
\newblock \doi{10.18653/v1/P16-1162}.
\newblock URL \url{https://aclanthology.org/P16-1162}.

\bibitem[Shridhar et~al.(2022)Shridhar, Manuelli, and Fox]{shridhar2022cliport}
Shridhar, M., Manuelli, L., and Fox, D.
\newblock Cliport: What and where pathways for robotic manipulation.
\newblock In \emph{Conference on Robot Learning}, pp.\  894--906. PMLR, 2022.

\bibitem[Silver et~al.(2016)Silver, Huang, Maddison, Guez, Sifre, Van
  Den~Driessche, Schrittwieser, Antonoglou, Panneershelvam, Lanctot,
  et~al.]{silver2016mastering}
Silver, D., Huang, A., Maddison, C.~J., Guez, A., Sifre, L., Van Den~Driessche,
  G., Schrittwieser, J., Antonoglou, I., Panneershelvam, V., Lanctot, M.,
  et~al.
\newblock Mastering the game of go with deep neural networks and tree search.
\newblock \emph{nature}, 529\penalty0 (7587):\penalty0 484--489, 2016.

\bibitem[Singh et~al.(2020)Singh, Liu, Zhou, Yu, Rhinehart, and
  Levine]{singh2020parrot}
Singh, A., Liu, H., Zhou, G., Yu, A., Rhinehart, N., and Levine, S.
\newblock Parrot: Data-driven behavioral priors for reinforcement learning.
\newblock \emph{arXiv preprint arXiv:2011.10024}, 2020.

\bibitem[Srivastava et~al.(2019)Srivastava, Shyam, Mutz, Ja{\'s}kowski, and
  Schmidhuber]{srivastava2019training}
Srivastava, R.~K., Shyam, P., Mutz, F., Ja{\'s}kowski, W., and Schmidhuber, J.
\newblock Training agents using upside-down reinforcement learning.
\newblock \emph{arXiv preprint arXiv:1912.02877}, 2019.

\bibitem[Su et~al.(2020)Su, Zhu, Cao, Li, Lu, Wei, and Dai]{su2020vlbert}
Su, W., Zhu, X., Cao, Y., Li, B., Lu, L., Wei, F., and Dai, J.
\newblock Vl-bert: Pre-training of generic visual-linguistic representations,
  2020.

\bibitem[Tirumala et~al.(2020)Tirumala, Galashov, Noh, Hasenclever, Pascanu,
  Schwarz, Desjardins, Czarnecki, Ahuja, Teh, et~al.]{tirumala2020behavior}
Tirumala, D., Galashov, A., Noh, H., Hasenclever, L., Pascanu, R., Schwarz, J.,
  Desjardins, G., Czarnecki, W.~M., Ahuja, A., Teh, Y.~W., et~al.
\newblock Behavior priors for efficient reinforcement learning.
\newblock \emph{arXiv preprint arXiv:2010.14274}, 2020.

\bibitem[Tsimpoukelli et~al.(2021)Tsimpoukelli, Menick, Cabi, Eslami, Vinyals,
  and Hill]{tsimpoukelli2021multimodal}
Tsimpoukelli, M., Menick, J., Cabi, S., Eslami, S. M.~A., Vinyals, O., and
  Hill, F.
\newblock Multimodal few-shot learning with frozen language models, 2021.

\bibitem[Vaswani et~al.(2017)Vaswani, Shazeer, Parmar, Uszkoreit, Jones, Gomez,
  Kaiser, and Polosukhin]{vaswani2017attention}
Vaswani, A., Shazeer, N., Parmar, N., Uszkoreit, J., Jones, L., Gomez, A.~N.,
  Kaiser, L., and Polosukhin, I.
\newblock Attention is all you need, 2017.

\bibitem[Vinyals et~al.(2019)Vinyals, Babuschkin, Czarnecki, Mathieu, Dudzik,
  Chung, Choi, Powell, Ewalds, Georgiev, et~al.]{vinyals2019grandmaster}
Vinyals, O., Babuschkin, I., Czarnecki, W.~M., Mathieu, M., Dudzik, A., Chung,
  J., Choi, D.~H., Powell, R., Ewalds, T., Georgiev, P., et~al.
\newblock Grandmaster level in starcraft ii using multi-agent reinforcement
  learning.
\newblock \emph{Nature}, 575\penalty0 (7782):\penalty0 350--354, 2019.

\bibitem[Wang et~al.(2018{\natexlab{a}})Wang, Singh, Michael, Hill, Levy, and
  Bowman]{wang2018glue}
Wang, A., Singh, A., Michael, J., Hill, F., Levy, O., and Bowman, S.~R.
\newblock Glue: A multi-task benchmark and analysis platform for natural
  language understanding.
\newblock \emph{arXiv preprint arXiv: Arxiv-1804.07461}, 2018{\natexlab{a}}.

\bibitem[Wang et~al.(2018{\natexlab{b}})Wang, Liao, Ba, and
  Fidler]{wang2018nervenet}
Wang, T., Liao, R., Ba, J., and Fidler, S.
\newblock Nervenet: Learning structured policy with graph neural networks.
\newblock In \emph{International Conference on Learning Representations},
  2018{\natexlab{b}}.

\bibitem[Watkins \& Dayan(1992)Watkins and Dayan]{watkins1992q}
Watkins, C.~J. and Dayan, P.
\newblock Q-learning.
\newblock \emph{Machine learning}, 8\penalty0 (3-4):\penalty0 279--292, 1992.

\bibitem[Wolf et~al.(2020)Wolf, Debut, Sanh, Chaumond, Delangue, Moi, Cistac,
  Rault, Louf, Funtowicz, Davison, Shleifer, von Platen, Ma, Jernite, Plu, Xu,
  Scao, Gugger, Drame, Lhoest, and Rush]{wolf2020huggingfaces}
Wolf, T., Debut, L., Sanh, V., Chaumond, J., Delangue, C., Moi, A., Cistac, P.,
  Rault, T., Louf, R., Funtowicz, M., Davison, J., Shleifer, S., von Platen,
  P., Ma, C., Jernite, Y., Plu, J., Xu, C., Scao, T.~L., Gugger, S., Drame, M.,
  Lhoest, Q., and Rush, A.~M.
\newblock Huggingface's transformers: State-of-the-art natural language
  processing, 2020.

\bibitem[Wu et~al.(2019)Wu, Tucker, and Nachum]{wu2019behavior}
Wu, Y., Tucker, G., and Nachum, O.
\newblock Behavior regularized offline reinforcement learning.
\newblock \emph{arXiv preprint arXiv:1911.11361}, 2019.

\bibitem[Yen-Chen et~al.(2020)Yen-Chen, Zeng, Song, Isola, and
  Lin]{yen2020learning}
Yen-Chen, L., Zeng, A., Song, S., Isola, P., and Lin, T.-Y.
\newblock Learning to see before learning to act: Visual pre-training for
  manipulation.
\newblock In \emph{2020 IEEE International Conference on Robotics and
  Automation (ICRA)}, pp.\  7286--7293. IEEE, 2020.

\bibitem[Yu et~al.(2020)Yu, Quillen, He, Julian, Hausman, Finn, and
  Levine]{yu2020meta}
Yu, T., Quillen, D., He, Z., Julian, R., Hausman, K., Finn, C., and Levine, S.
\newblock Meta-world: A benchmark and evaluation for multi-task and meta
  reinforcement learning.
\newblock In \emph{Conference on Robot Learning}, pp.\  1094--1100. PMLR, 2020.

\bibitem[Zeng et~al.(2020)Zeng, Florence, Tompson, Welker, Chien, Attarian,
  Armstrong, Krasin, Duong, Sindhwani, et~al.]{zeng2020transporter}
Zeng, A., Florence, P., Tompson, J., Welker, S., Chien, J., Attarian, M.,
  Armstrong, T., Krasin, I., Duong, D., Sindhwani, V., et~al.
\newblock Transporter networks: Rearranging the visual world for robotic
  manipulation.
\newblock \emph{arXiv preprint arXiv:2010.14406}, 2020.

\bibitem[Zhu et~al.(2020)Zhu, Wong, Mandlekar, and
  Mart{\'\i}n-Mart{\'\i}n]{zhu2020robosuite}
Zhu, Y., Wong, J., Mandlekar, A., and Mart{\'\i}n-Mart{\'\i}n, R.
\newblock robosuite: A modular simulation framework and benchmark for robot
  learning.
\newblock \emph{arXiv preprint arXiv:2009.12293}, 2020.

\end{thebibliography}
\bibliographystyle{icml2021}
\onecolumn
\appendix
\section{Appendix}
\subsection{Hyperparameters \& Training Details}
\begin{table}[!htb]
    \centering
    \begin{tabular}{l|c}
        \toprule
        Hyperparameter & Value \\
        \midrule
         \# Layers & 3  \\
         \# Attention Heads& 1 \\
         Activation fn. & ReLU \\
         Batch size & 64 \\
         Context & 20 \\
         Return-to-go conditioning & 6000 HalfCheetah \\ 
         & 3600 Hopper \\
         & 5000 Walker \\
         Dropout & 0.2 \\
         Learning rate & $1$e-$4$ \\
         LR Warmup & 5000 steps \\
         \bottomrule
    \end{tabular}
    \caption{Hyperparameters used for OpenAI Gym}
    \label{tab:my_label}
\end{table}
\paragraph{Implementation details} Pre-trained models are trained with and taken from the HuggingFace Transformers library \citep{wolf2020huggingfaces}. The model code for our GPT2 model is \texttt{gpt2}, CLIP is \texttt{openai/clip-vit-base-patch32}, and iGPT \texttt{openai/imagegpt-small}.
\begin{table}[!htb]
    \centering
    \begin{tabular}{c|cc}
    \toprule
\bf Model & \bf Parameter Count & \bf Num. Tokens \\ \midrule
        DT & 596K & --- \\
        \chibit & 596K & $10^7$\\
        iGPT  &  84M & $10^{10}$ \\
        GPT-2 & 84M & $10^{10}$\\
        CLIP & 38M & $10^{10}$ \\\bottomrule
    \end{tabular}
    \caption{Model parameter counts and number of unique pre-training tokens}
    \label{tab:my_label}
\end{table}
\paragraph{Language Model Pre-training with larger sizes} For our large sized pre-trained models in our model scale experiments, we use the following dimensions:
\begin{table}[!htb]
    \centering
    \begin{tabular}{ccccc}
    \toprule
        \bf Param. Count & Model Dim. & Num. Heads & Num. Layers  \\
        \midrule
         3M & 256 & 4 & 4 \\
         18M & 512 & 8 & 6 \\
         84M & 768 & 12 & 12 \\
        \bottomrule
    \end{tabular}
    \caption{Parameter count for various pre-trained models used in our model scale experiments.}
    \label{tab:my_label}
\end{table}

\section{Attention Visualization}
We visualize the attention weights with a temperature of $0.1$ to improve visual interpretation.
\newpage
\section{Reproduction of DT results versus DT results in \citet{chen2021decision}}
We re-run the results in \citet{chen2021decision} and include them for reference in Table~\ref{tab:gg}.
\begin{table*}[!htb]
\centering
\footnotesize
\resizebox{!}{!}{\begin{tabular}{llrr}
\toprule
\multicolumn{1}{c}{\bf Dataset} &  \multicolumn{1}{c}{\bf Environment} &  \multicolumn{1}{c}{\bf DT} & \multicolumn{1}{c}{\bf DT(ours)} \\
\midrule
\multirowcell{3}{Medium Expert} & HalfCheetah &$86.8\pm1.3$ & $86.5\pm0.8$ \\
& Hopper      &$107.6\pm1.8$ & $107.4\pm2.0$\\
& Walker      &$ 108.1\pm0.2$ & $108.4\pm0.1$\\
\midrule
\multirowcell{3}{Medium} & HalfCheetah &$42.6\pm0.1$ & $42.1\pm0.3$ \\
& Hopper      &$67.6\pm1.0$ & $68.1\pm3.1$\\
& Walker      &$ 74.0\pm1.4$ & $74.4\pm1.9$\\
\midrule
\multirowcell{3}{Medium Replay} & HalfCheetah &$36.6\pm0.8$ & $36.2\pm1.4$ \\
& Hopper      &$82.7\pm7.0$ & $80.4\pm6.3$ \\
& Walker      &$ 66.6\pm3.0$ & $67.0\pm2.4$ \\
\midrule
\multicolumn{2}{c}{\bf Average (All Settings)}    & 74.7  & 74.5 \\
\bottomrule
\end{tabular}}
\caption{Re-implementation of Decision Transformer using their codebase\footnote{https://github.com/kzl/decision-transformer}}
\label{tab:gg}
\end{table*}

\section{Performance profiles}
We compute statistical significance tests using \texttt{rliable} \citep{DBLP:conf/nips/AgarwalSCCB21} on OpenAI Gym. Specifically, as we are only comparing two algorithms DT \citep{chen2021decision} and ChibiT, we only plot performance profiles and the boostrapped confidence interval measure.
\begin{figure*}[!htb]
	\centering
	\includegraphics[width=0.95\textwidth]{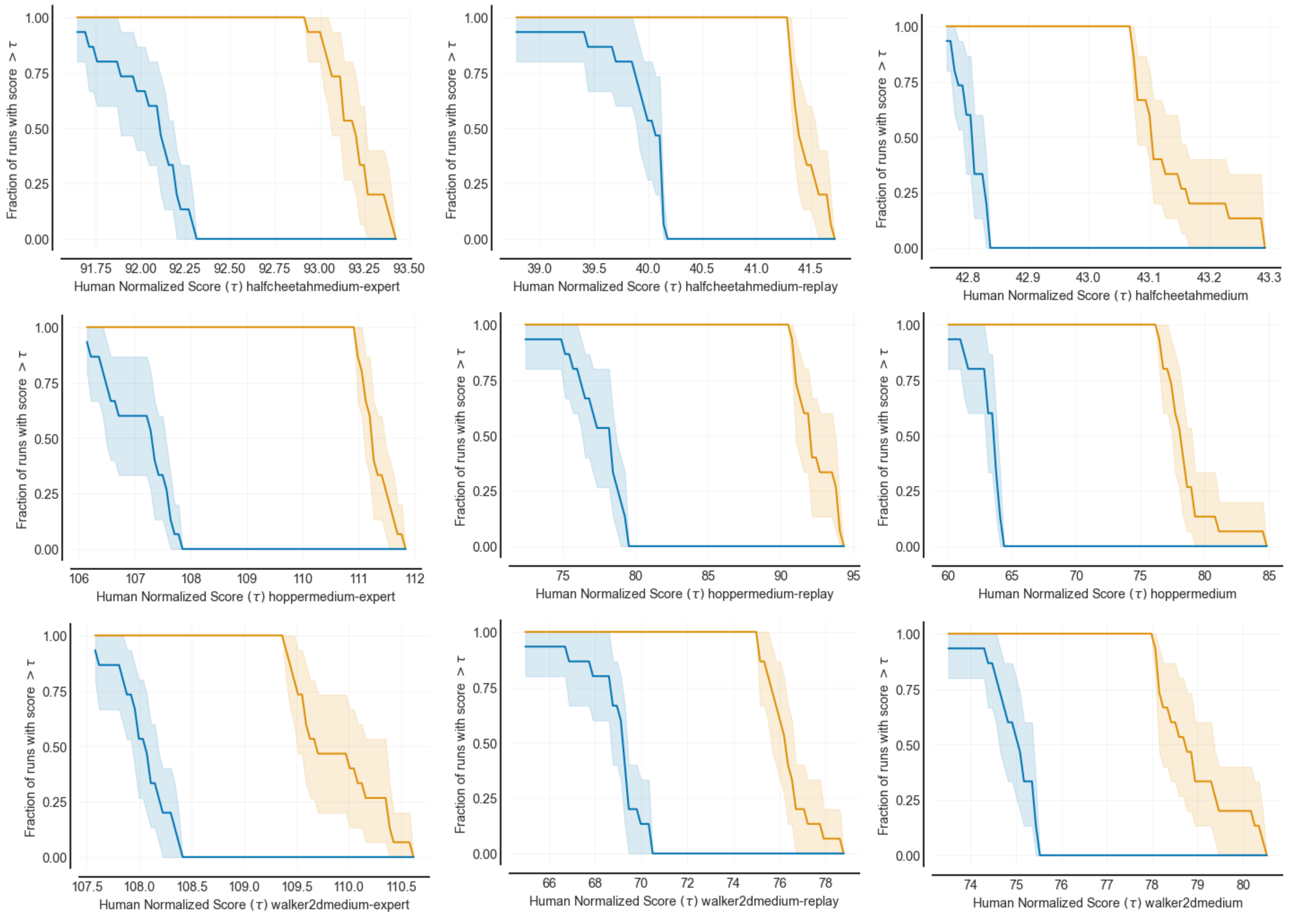} \label{fig:perf_profile}
	\caption{\textbf{Performance profiles on D4RL datasets.} Yellow colors represent ChibiT and blue colors represent Decision Transformer (DT). 
	We report the profiles based on score distributions over 10 runs using different random seeds.
	Language model pre-trained models are consistently better than DT.
	}
\end{figure*}

\begin{figure}
    \centering
    \includegraphics[width=0.95\textwidth]{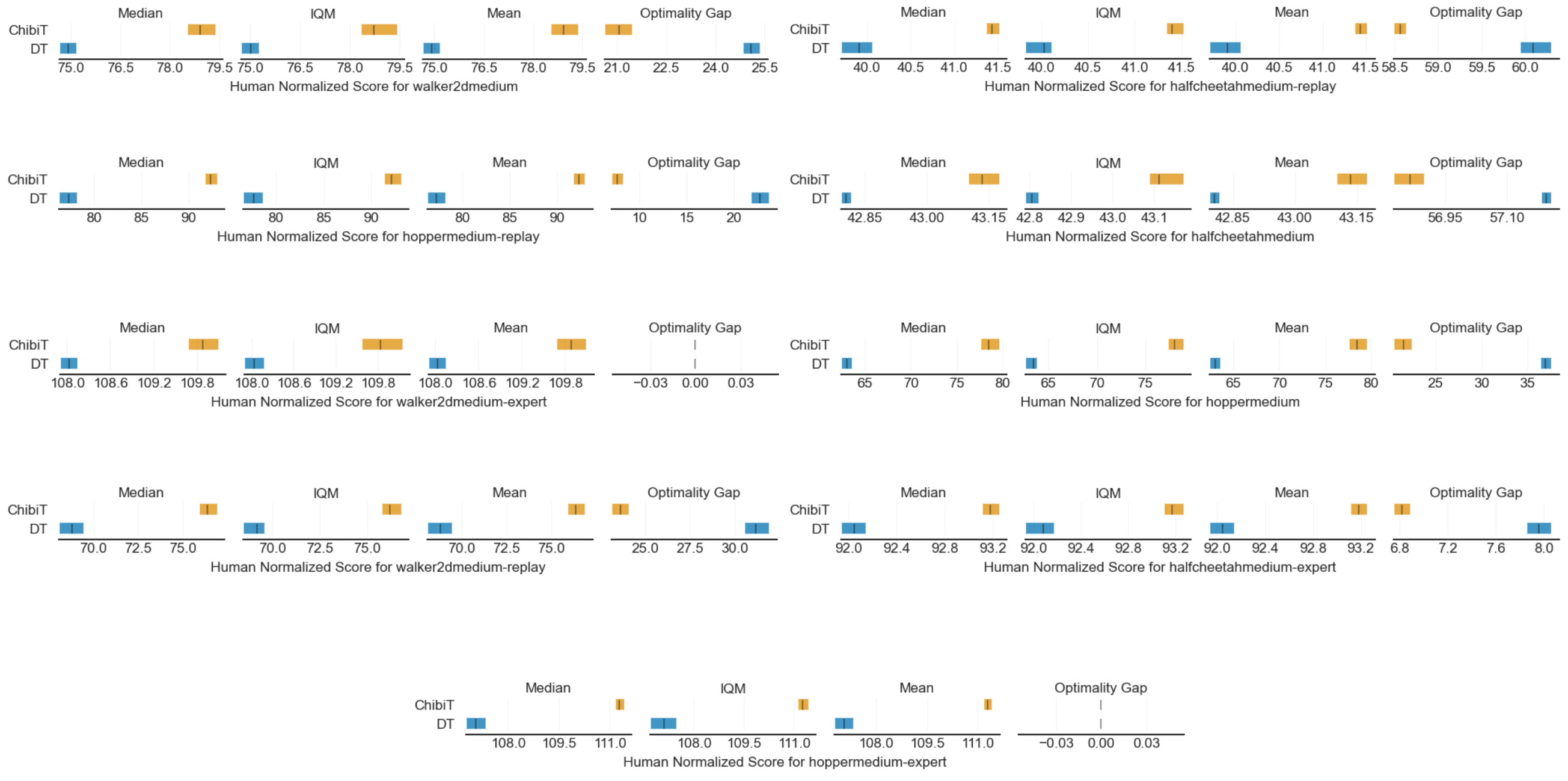}
    \caption{Bootstrapped confidence intervals (CIs) on D4RL datasets.Yellow colors represent ChibiT and blue colors represent Decision Transformer (DT). 
	We report the intervals based on score distributions over 10 runs using different random seeds.
	Language model pre-trained models are consistently better than DT.}
    \label{fig:boostrapped_ci}
\end{figure}

\end{document}